\documentclass[12pt]{article}   	
\usepackage[right=1in,left=1in,top=1in,bottom=1in]{geometry}                		
\usepackage{mathrsfs}
\usepackage{amssymb}
\usepackage{amsmath}
\usepackage{verbatim}
\usepackage{graphicx}
\usepackage{tikz}
\usepackage{setspace}

\title{Surreal Decisions }

\author{Eddy Keming Chen\thanks{Department of Philosophy,  University of California San Diego, 9500 Gilman Dr, La Jolla, CA 92093-0119. Website: www.eddykemingchen.net. Email: eddykemingchen@ucsd.edu} \text{} and Daniel Rubio\thanks{University Center for Human Values, Princeton University, 5 Ivy Lane, Princeton, NJ 08544. Website: www.danielkfrubio.com. Email: danielkfrubio@gmail.com}}

\date{\textit{Philosophy and Phenomenological Research}, 2020, 100(1): 54-74}

\begin{document}
\maketitle
\begin{abstract}
\noindent
Although expected utility theory has proven a fruitful and elegant theory in the finite realm, attempts to generalize it to infinite values have resulted in many paradoxes. In this paper, we argue that the use of John Conway's surreal numbers shall provide a firm mathematical foundation for transfinite decision theory. To that end, we prove a surreal representation theorem and show that our surreal decision theory respects dominance reasoning even in the case of infinite values. We then bring our theory to bear on one of the more venerable decision problems in the literature: Pascal's Wager. Analyzing the wager showcases our theory's virtues and advantages. To that end, we analyze two objections against the wager: Mixed Strategies and Many Gods. After formulating the two objections in the framework of surreal utilities and probabilities, our theory correctly predicts that (1) the pure Pascalian strategy beats all mixed strategies, and (2) what one should do in a Pascalian decision problem depends on what one's credence function is like. Our analysis therefore suggests that although Pascal's Wager is mathematically coherent, it does not deliver what it purports to, a rationally compelling argument that people should lead a religious life regardless of how confident they are in theism and its alternatives. 

\end{abstract}

\vspace{10pt}

\hspace*{3,6mm}\textit{Keywords:  decision theory, dominance principles, infinite values, surreal numbers, Pascal’s Wager, mixed strategies, representation theorem, pragmatic argument}   

\newpage

\begingroup
\singlespacing
\tableofcontents
\endgroup

\section{Introduction}

Infinities have bedeviled decision theorists since the early days of the Port Royal Logic. First came Pascal's Wager, with the use of single-state infinite utilities underlying the (in)famous argument for theism; next came the St. Petersburg Game---the first game to reveal the counterintuitive consequences of infinitely many possible outcomes. Many others have followed. Since then, decision theorists have taken two broad approaches: the \textit{strict finitist strategy}, which steadfastly bans infinities outright, and the \textit{open arms strategy}, which has embraced infinity in all of its counterintuitive glory and employed increasingly sophisticated mathematics to smooth out the paradoxes. 

We consider the strict finitist strategy---despite its impressive pedigree---to be less than fully satisfactory.\footnote{In fact, it is not satisfactory at all when the agent has genuinely non-Archimedean preferences: for three lotteries $x$,  $y$, and $z$, the agent would not trade $y$ for any real-valued multiples of $x$, and she would not trade $z$ for any real-valued multiples of $y$.} We find decision problems with transfinite utilities (or transfinite state spaces) to be no less sensible or well-formed than those involving only finite ones. Unfortunately, standard expected utility (EU) theory is not well-equipped to handle them.  After examining the structure of infinitistic decision puzzles, we observe that the origin of the problem lies not in the general conceptual framework of EU theory but in the limitation of  \emph{the mathematical representations of infinite utilities}. That is, in the infinite realm, the usual mathematical framework (including real analysis) no longer faithfully represents the rational EU preference structures. In Alan H\'ajek's terms, we need better technology! We therefore offer, as a conservative extension of finitist EU theory and a solution to this problem, a decision theory framed entirely in John Conway's surreal numbers. In this work, we confine ourselves to finite state spaces; while we believe that future work will show surreal decision theory to be a fruitful framework for approaching the problems of infinite state spaces, this is still work in progress and will depend on development of ongoing research in surreal analysis. 

Our paper proceeds as follows: In \S 2, we  offer some motivations for introducing surreals into decision theory. In \S 3, we  give the reader a brief guided tour of surreal mathematics, highlighting their most useful properties,  then prove a von Neumann-Morgenstern representation theorem for surreal utilities, and finally discuss the meaning of surreal credences. In \S 4, we apply our theory to bear on Pascal's Wager and two oft-cited objections against it: Mixed Strategies and Many Gods. After formulating the two objections in the framework of surreal utilities and probabilities, our theory suggests that although Pascal's Wager is mathematically coherent, it does not deliver what it purports to do, i.e. convince people that they should lead a Christian life regardless of how confident they are in theism and its alternatives. Next, we will explore an additional wrinkle: many popular religions do not offer a one-size-fits-all afterlife; in some forms of Buddhism, Christianity,  Islam, and Judaism, we find a `degrees of glory' eschatology, in which what sort of good afterlife is offered to the faithful (or bad afterlife to the infidel) depends on their earthly deeds. As we shall see, it can be faithfully modeled in our theory.

\section{Problems With the Infinite}

According to the standard expected utility (EU) theory, we compute the value of a gamble by taking its expected utility. Roughly, this means that we multiply the utility of each possible outcome by the utility an agent assigns to that outcome. More formally, where G is a gamble \{$x_1, cr_1;x_2, cr_2;...;x_n, cr_n;...\}$, $cr$ is the agent's credence function, and $u(x_i)$ is the agent's utility for state $x_i$, the expected utility of the gamble G is:  

$$EU(G) = \sum\limits_{i=1}^n cr_iu(x_i)$$
\noindent
where $\sum\limits_{i=1}^n cr_i=1$.

In orthodox decision theory, it is assumed that utilities are bounded and draw numbers from $\mathbb{R}$. This stipulation leads to an elegant theory, but there are a number of gambles that seem possible but that violate the boundedness of utility constraint. Pascal's Wager and the St. Petersburg Game are the most historically noteworthy. In order to analyze these, many decision theorists have considered a simple infinitistic decision theory where the domain of the utility function is enriched with a positive and a negative infinity and the expected utility rule is modified to accommodate countable sums.  

But this simple change, together with the prescription that agents should (only) maximize expected utilities, yields counter-intuitive advice in many situations. Indeed, it yields advice that violates the traditional Independence axiom of standard decision theories.  Consider the following trio of simple gambles:
\begin{quote}
\textit{Infinity or Nothing}: you are offered a coin flip that yields infinite utility if heads, and nothing if tails. The gamble is thus $G_1$ = \{$.5, \infty; .5, 0$\}.\\
\\
\textit{Infinity or Something}: you are offered a coin flip that yields infinite utility if heads, and utility 10,000 if tails. The gamble is thus $G_2$ = \{$.5, \infty; .5, 10,000$\}.\\
\\
\textit{Infinity or Bust}:  you are offered a coin flip that yields infinite utility if heads, and -10,000 utility if tails. The gamble is thus $G_3$ = \{$.5, \infty; .5, -10,000$\}.
\end{quote}
We submit that the rational preferences over $G_1$, $G_2$, and $G_3$ is: $G_2 > G_1 > G_3$. Dominance reasoning agrees with us, since $G_2$ weakly dominates both $G_1$ and $G_3$, while $G_1$ weakly dominates $G_3$. But (as the reader can easily verify) $EU(G_1)=EU(G_2)=EU(G_3)=\infty$, and so standard EU theory prescribes indifference. 

And while EU makes bad predictions in the first simple series, matters get worse with the next trio of gambles:
\begin{quote}
\textit{Fair Infinity}: you are offered a coin flip that yields infinite utility if heads, and infinite disutility if tails. The gamble is thus $G_4$ = \{$.5, \infty; .5, -\infty$\}.\\
\\
\textit{Biased Positive Infinity}: you are offered a coin flip that yields infinite utility if heads, and infinite disutility if tails. the coin is biased 9:1 in favor of heads. The gamble is thus $G_5$ = \{$.9, \infty; .1, -\infty$\}.\\
\\
\textit{Biased Negative Infinity}:  you are offered a coin flip that yields infinite utility if heads, and infinite disutility if tails.  The coin is biased 9:1 against heads. The gamble is thus $G_6$ = \{$.1, \infty; .9, -\infty$\}.
\end{quote}
As before, we contend that there is a clear order of rational preference over these gambles: $G_5 > G_4 > G_6$. But EU theory makes no predictions here, for the value of $\infty-\infty$ is not well-defined.

Part of the problem comes from the inherent vagueness of $\infty$. While it is good enough for the calculus or real analysis, any rigorous attempt to do transfinite arithmetic must enter Cantor's paradise. Thus, a first-pass fix would assign cardinal numbers as utilities and use cardinal arithmetic to calculate the expected utilities. Using cardinal arithmetic, we would get the following values for our simple gambles:
\begin{itemize}
\item [$G_1$] = $.5\aleph_0$
\item [$G_2$]= $.5\aleph_0+5,000$
\item [$G_3$]= $.5\aleph_0-5,000$
\item [$G_4$]= $.5\aleph_0-.5\aleph_0$
\item [$G_5$]= $.9\aleph_0-.1\aleph_0$
\item [$G_6$]= $.1\aleph_0-.9\aleph_0$
\end{itemize}
Unfortunately, this is of no help to us. Cardinal arithmetic also has the absorption property. Assuming the axiom of choice, if either $\kappa$ or $\mu$ is infinite, then $\kappa+\mu=max\{\kappa, \mu\}$, and $\kappa\times\mu=max\{\kappa,\mu\}$. Furthermore, $\aleph_0-\aleph_0$ is  not well-defined. So the results  still diverge from the usual intuitions (or the advice of dominance reasoning), and the latter series is still without a numerical evaluation.

Cardinal arithmetic is not the only way to operate on transfinite numbers. There is also ordinal arithmetic. Ordinal arithmetic lacks the absorption property, but is non-commutative, which makes it singularly unfit for decision theory. And to add insult to injury, $\omega-\omega$ remains undefined. 

\section{A Surreal Solution}

As we saw in \S 2, transfinite decision theory requires the ability to perform arithmetic operations on finite and infinite numbers, with commutativity and non-absorption (and other standard desirable properties of addition), and where every number---finite and transfinite---has an additive inverse (so that, for example, $\omega-\omega$, is defined). More precisely, we require:
\begin{enumerate}
\item an ordered-field including all reals and ordinals;
\item addition in that field that is commutative, non-absorptive, and such that each element has an additive inverse;
\item multiplication in that field that is commutative, non-absorptive, and such that each non-zero element has a multiplicative inverse.
\end{enumerate}
In short: a number system and accompanying operations that allow us to treat finite and transfinite numbers in similar and familiar ways.

Fortunately, John Conway discovered (or invented, depending on your philosophy of mathematics) such a field, and began its exploration in his \textit{On Numbers and Games} (1974). Conway called the objects he discovered \textit{surreal} numbers. For those familiar with Dedekind's construction of the reals out of the rationals, it may be helpful to note that Conway's construction is quite similar to Dedekind's. Except, rather than using the rationals, Conway uses the ordinals. Nevertheless, we can think of surreal numbers as (equivalence classes of) ``Dedekind cuts'' on ordinals. They are defined recursively as follows:\footnote{Conway [1974]}\\
\\
{\sc definition 1:} If $L$ and $R$ are sets of numbers, and no $x\in L \ge$ any $y\in R$, then $\{L|R\}$ is a number.\\ 
\\
{\sc convention 1}: If x=$\{L|R\}$, we will write $x^L$ as a convention for the typical member of L, and $x^R$ for the typical member of R\\
\\
{\sc definition 2}: $x\ge y$ iff no $x^R\le y$ and no $y^L\ge x$\\
\\
Other familiar ordering relations are defined in the usual way.\footnote{$x\not\ge y$ iff not $x\ge y, x > y$ iff $x\ge y$ and $y\not\ge x, x=y$ iff $x\ge y$ and $y\ge x$. } 

Definition 1 looks circular. Fortunately, the null set is trivially a set of numbers, and so our first surreal number is \{$\varnothing|\varnothing\}=0$.\footnote{A similar trick saves Definition 2 from circularity, for it allows us to prove that $0\ge 0$.} From $0$, we gain two new numbers: \{$0|\varnothing\} = 1$ and \{$\varnothing|0\}=-1$. From these numbers, we can find yet more numbers, including \{$\{0, 1, 2, 3, 4 ... \} |\varnothing\}=\omega$, \{$\{0\} |\{1, 1/2, 1/4, 1/8 ...\} \}=1/ \omega$, \{$\{0, 1, 2, 3, 4 ... \} |\omega \}=\omega - 1$. In order to avoid tedious iterations, we can see the structure of the surreals laid out in Figure 1. We use \textbf{No} to denote the class of numbers created by repeated application of definition 1, and the iteration of definition 1 on which $n$ is found its `birthday.'\footnote{So the birthday of $0$ is day $0$ the birthday of $1, -1$ is day $1$, etc. Notice that Definitions 1 and 2 also explain why surreal numbers are equivalence classes of ``cuts.'' For example, although $2\equiv \{ 1 | \}$, i.e. the name of $\{ 1 | \}$ is ``two,'' we can easily prove that $2= \{ -1 , 0, 1 | \} = \{-1, 1 | \} =\{ 0,1 | \} =\{ 1 | \}$.}

\begin{figure}
\centering
\includegraphics[height = 10 cm]{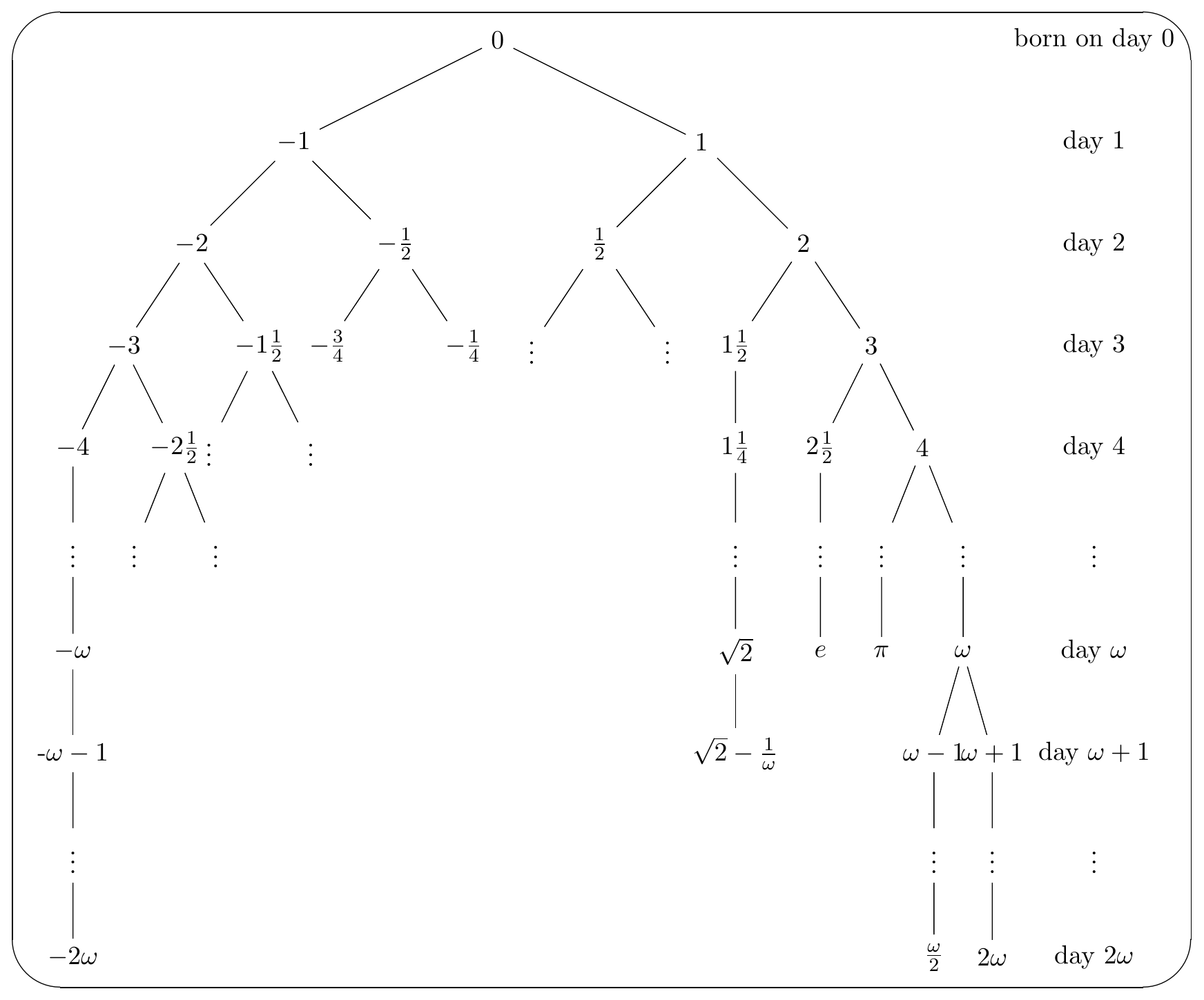}
\caption{The Surreal Tree}
\end{figure}
With a hearty stock of numbers, we can now set about defining arithmetic operations.\\
\\
{\sc definition 3}: x + y = \{$x^L+y, x+ y^L|x^R+y, y^R+x\}$\footnote{Again, this definition looks circular, as we seem to be defining addition from addition. Here we used the notational convention that adding a number to a set is just adding that number to every member of the set. For example, $x^{L}+y$ means adding $y$ to every member of $x^L$. Since there is no member in the empty set $\emptyset$, we automatically have the base case. Applying the base case, we have 
$$0+0 = \{ \emptyset + 0, 0 + \emptyset | \emptyset + 0, 0 + \emptyset  \}= \{ \emptyset | \emptyset \} = 0;$$ 
$$0+1 = \{ \emptyset + 1, 0 + 0 | \emptyset + 1, 0 + \emptyset  \}= \{ 0 |  \} = 1.$$
(In the second to last step, we have applied a theorem that allows us to simplify the surreal numbers without losing the equality.)  Definition 5 secures the base case in the same way.}\\
\\
{\sc definition 4}: -x = \{-$x^L|$-$x^R$\}\\
\\
{\sc definition 5}: $x\times y = \{x^L\times y+y^L\times x-x^L\times y^L, x^R\times y+y^R\times x-x^R\times y^R|x^L\times y+y^R\times x-x^L\times y^R, x^R\times y+y^L\times x-x^R\times y^L\}$\\
\\
These definitions make \textbf{No} an ordered field including all reals and all ordinals (in fact, Conway proves that it is a universally embedding field). We refer the interested reader to Conway for the proofs and further details.\footnote{Conway [1974], 15-44}

We end this quick introduction to the surreals with some remarks about the status of real numbers. As we can see from the surreal tree (Figure 1), we can find \emph{bona fide} real numbers such as $ -2\frac{1}{2}, -2, -1, 0, \frac{1}{3}, 1, \sqrt{2}, 2, \pi$ among the surreals. Moreover, all of them can be explicitly identified. For example, with a natural mapping function $f$, we can identify the real number $0$ with $\{|\} = f(0)$, any positive integer with $\{ f (x-1) | \}$,  any negative integer with $\{ | f (x+1) \}$, and any rational that is a dyadic fraction $\frac{j}{2^{k}}$ (meaning that $j, k$ are integers and $k$ is positive) with $\{  f(\frac{j-1}{2^{k}}) | f( \frac{j+1}{2^{k}} ) \}$. For any real number $x$ (such as $\frac{1}{3}$ and $\pi$) that is not a dyadic fraction, we define it as  $\{L | R \}$, where $L$ is the set of all dyadic fractions $\frac{j}{2^{k}}$  smaller than $x$ and $R$ is the set of all dyadic fractions larger than $x$.   Two examples: (1) $\frac{1}{3} = \{A | B \}$, where $A$ is the set of all dyadic fractions $\frac{j}{2^{k}}$  smaller than $\frac{1}{3}$, and $B$ is the set of all dyadic fractions larger than $\frac{1}{3}$; (2) $\pi = \{C | D \}$, where $C$ is the set of all dyadic fractions smaller than $\pi$, and $D$ is the set of all dyadic fractions larger than $\pi$. Moreover, the mapping function $f$ is provably a faithful embedding from the reals into the surreals (a homomorphism that preserves ordering, as well as addition, multiplication, and other algebraic operations).\footnote{We refer the interested reader to Gonshor [1986] for the mathematical details and Chris Tondering's excellent notes for a clear introduction to the basic ideas.} 


\subsection{A Surreal Representation Theorem}

Given the above properties of surreal numbers, can we model rational preference structures using the surreal field? Indeed we can. Here we prove that we can represent a rational agent's preferences by a surreal-valued utility function. In so doing, we establish a mathematical foundation for surreal decision theory.\\

\noindent
{\sc Notation 1}:  Let  $\star$ denote the natural embedding from the standard universe into the surreal universe. Let  \textbf{No} denote a surreal model. \\

\noindent
{\sc theorem 1 (Surreal von Neumann-Morgenstern Theorem)}: Let X be a finite space of lotteries, and let $\preceq$ be a binary relation $\subseteq X \times X$. Then $\preceq$ admits an expected utility representation $U : X \rightarrow \textbf{No}$ such that $\forall x, y \in X$:

$U(x) \leq U(y) \Leftrightarrow x \preceq y$ if and only if $\preceq$ satisfies all of the following:
\begin{enumerate}
\item Completeness: $\forall x, y \in X,$ either $x \preceq y$ or $y \preceq x$.
\item Transitivity: $\forall x, y, z \in X,$ if $x \preceq y$ and $y \preceq z$, then $x \preceq z$.
\item Continuity$\star$: $\forall x, y, z \in X,$ if $x \preceq y \preceq z$, then there exist a surreal $p \in \star[0, 1]$ such that $px + (1 - p)z \sim y$.
\item Independence$\star$: $\forall x, y, z \in X, \forall p \in \star(0, 1],$ $x \preceq y$ if and only if $px + (1 - p)z \preceq py + (1 - p)z$.
\end{enumerate}

\textit{Proof:} See Appendix.

\subsection{A Simple Application}
With surreal arithmetic thus defined and the representation theorem proven, we can represent a rational agent's preferences by a surreal-valued utility function. We now return to the games of \S 2. Recall the calculations we needed to make (with $\infty$ precisified as the ordinal $\omega$):
\begin{itemize}
\item [$G_1$] = $.5\omega$
\item [$G_2$]= $.5\omega+5,000$
\item [$G_3$]= $.5\omega-5,000$
\item [$G_4$]= $.5\omega-.5\omega$
\item [$G_5$]= $.9\omega-.1\omega$
\item [$G_6$]= $.1\omega-.9\omega$
\end{itemize}
With either ordinal or cardinal arithmetics, we get results that disobey dominance reasoning and are therefore unacceptable. However, with surreal arithmetic operations, we get the intuitive results that respect dominance principles. $G_2>G_1>G_3$, and $G_5>G_4>G_5$.

Detailed calculations, even in this simple example, can be tedious. But for an illustration, we will show (with the help of some theorems in surreal analysis) that $G_2>G_1$. (This method can be generalized to prove a general theorem about dominance in surreal decision theory.) We can use the definitions to show: 
$$.5\omega = .5\omega$$
$$0 \leq 5,000$$
$$.5\omega + 0 = .5\omega$$
Moreover, it is a theorem that $x< x' \wedge y< y' \Rightarrow x+y < x' + y'$. So we can conclude that $.5\omega < .5\omega + 5,000$. 

It is worth noting, although it plays no part here, that $\omega-\omega$ is defined, and is 0 (because -$\omega$ is the additive inverse of $\omega$). 

\subsection{Surreal Credences}

So far, especially in the statement of our representation theorem, we have assumed that credences can come in surreal values. Here, we give some reasons for thinking that this is a plausible assumption. Our goal here is  to argue that surreal credences are viable -- that is, that surreal numbers can be used to model mental states without falling into incoherence. We do not claim, however, that they are the only or the best tool for the job. We do this in two parts. First: we argue that we can understand what it means to have a surreal credence by considering cases where surreal mathematics provides a natural  model of certain intuitively rational preferences. Second, we argue that recent work in non-Archimedean probability theory makes surreal mathematics a viable model for probability on finite state spaces.\footnote{We thank an anonymous referee for suggesting that we make these clear.}

\subsubsection{What Surreal Credences Might Represent}

We are used to probabilities being Archimedean, represented by real numbers. (Some people  think that even  real numbers are too fine-grained to represent credences.) Indeed, in our ordinary choice situations, nothing really calls for non-Archimedean credences that motivate representation by surreal numbers. However, since we would like to discuss Pascal's Wager, our situation will be somewhat out of the ordinary contexts (and Pascal insists that his Wager is a decision problem that all of us face). In the following, we shall consider two intuitively rational preferences that seem to lead to non-Archimedean credences.\footnote{Here we are granting the assumption that preferences are prior to credences. But see Eriksson \& H\'ajek (2007) for some interesting arguments that credences can be taken as more fundamental than preferences. This is also natural in the von Neumann-Morgenstern framework where probabilities (objective chances) are assumed. If we allow the possibility that credences are prior to betting preferences, then we can make sense of surreal credences in another way--via surreal chances and the Principal Principle. If there is a fair lottery with $\omega_0$ tickets, then it is plausible that the chance of any ticket being the winning one is $\frac{1}{\omega_0}$.} (Let us stipulate that money is linear in dollars and money is real-valued.)

Case 1. Rene is offered a ticket (that costs nothing and that pays \$10 in reward if it wins) in a fair countable lottery. She strictly prefers taking it to rejecting, but she will not pay for it. Since she strictly prefers it to the status quo, the ticket has positive expected utility (EU). So the credence of getting a payout is greater than 0. But since there's no amount of money she'll pay for it, $\forall x\in \mathbb{R},$ $0< EU (ticket) < x$. Therefore, Rene's credence that the ticket will win is a positive infinitesimal number.

Case 2. Blaise has a ticket for salvation that costs nothing and that pays a high reward if it wins. At the initial time $t_1$, there's no monetary price at which he will sell his ticket. So Blaise's expected utility of the ticket $EU_{t_1} (ticket)$ is infinite. But you give him tons of evidence against the ticket being genuine, such that the evidence at some later time $t_2$  will convince him to sell it at a price but still not give it away. Therefore, $\exists x \in \mathbb{R},$ $0< EU_{t_2} (ticket) \leq x$. Therefore, Blaise's posterior credence that the ticket is genuine is a positive infinitesimal number. 

We judge these preferences to be intuitively rational. Now, both cases lead to non-Archimedean credences. They can be modeled by surreal numbers, which include not only  infinities but also  infinitesimals that satisfy the multiplication properties used in these cases. Therefore, we can capture these preferences with surreal credences. 

Caveat: in both cases we make use of infinite state spaces. They might appear illegitimate  since we are in the context of justifying surreal decision theory for finite state spaces. However, the appearances are misleading. If infinitesimal credences are coherent in infinite state spaces, they should remain coherent in finite state spaces. We can, for example, appeal to Case 1 and Case 2 as hypothetical situations, which can provide interpretations of the surreal credences in Pascal's Wager and the final two axioms in the Surreal von Neumann-Morgenstern Theorem.\footnote{We should add that the technical questions about how to do surreal infinite sum (especially on conditionally convergent series)  do not come up in these two cases, as we are merely doing pair-wise comparisons for infinitely many pairs, not summing over infinitely many values. They are conceptually related but mathematically different.}


\subsubsection{Technical Prospects for a Surreal Probability Theory}

We are used to the standard classical probability theory, axiomatized by Kolmogorov. It is attractive for its simplicity and fruitfulness. 
Take, for example, the presentation of the Kolmogorov axioms from Benci et al. [2016]:

\begin{enumerate}
\item[K0] Domain and Range: The events are the elements of a $\sigma-$algebra $F\subset \mathscr{P}(\Omega)$ and the probability function is a function $P : F \rightarrow \mathbb{R}$.
\item[K1] Non-negativity: $\forall A \in F$, $P(A) \geq 0$.
\item[K2] Normalization: $P(\Omega) = 1$.
\item[K3] Additivity: If $A$ and $B$ are events and $A \bigcap B = \emptyset$, then $P(A\bigcap B) = P(A) + P(B)$. 
\item[K4] Continuity: Let $A = \bigcup_{n\in \mathbb{N}} A_n$, where $A_n \subset A_{n+1}$ are elements of $F$, then $P(A) = \lim_{n\rightarrow \infty} P(A_{n})$. 
\end{enumerate}
Notice that adding [K4] to [K0]--[K3] is equivalent to requiring Countable Additivity. 

Despite its simplicity and conceptual attractiveness, it cannot handle, for example, a fair lottery on  natural numbers (Wenmackers and Horsten [2013]). We have to  reject either [K2] Normalization or [K4] Continuity. It is natural to assign unity to the probability of the whole space, so it is reasonable to reject or modify [K4], which is equivalent to modify Countable Additivity. 

But what happens when we take away Countable Additivity? It has interesting mathematical and philosophical consequences. Benci et al. [2016] suggest an axiomatization of a finitely-additive probability theory with the additional assumption of Regularity:
\begin{enumerate}
\item[NAP0] Domain and Range: The events are all the subsets of  $\mathscr{P}(\Omega)$, which can be a finite or infinite sample space. The probability function is a total function $P : \mathscr{P}(\Omega) \rightarrow \mathscr{R}$, where $\mathscr{P}(\Omega)$ is the powerset of $\Omega$ and $\mathscr{R}$ is a superreal field (that is, an ordered field that contains the real numbers as a subfield). 
\item[NAP1] Regularity: $P(\emptyset) = 0$ and $\forall A \in \mathscr{P}(\Omega) \backslash \{\emptyset\}$, $P(A) > 0$.
\item[NAP2] Normalization: $P(\Omega) = 1$.
\item[NAP3] Additivity: If $A$ and $B$ are events and $A \bigcap B = \emptyset$, then $P(A\bigcap B) = P(A) + P(B)$. 
\end{enumerate}
By stipulation [NAP0], such a finitely additive probability function can have values in a non-Archimedean superreal field, of which the surreal field \textbf{No} is an instance. Moreover, they prove that if [NAP0]--[NAP3] hold, then: 

(1) $\forall A \in \mathscr{P}(\Omega), P(A) \in [0,1]_{\mathscr{R}}$;  

(2) $P(A) = 1 \leftrightarrow A=\Omega$; 

(3) If $\Omega$ is uncountable or $\Omega$ is countable and the lottery is fair ($\forall \omega, \tau \in \Omega, P(\omega) = P({\tau})$), then $\mathscr{R}$ is a non-Archimedean field. 

That is, if the probability function is defined an uncountable sample space or assigns ``fair lotteries,''  then the value field has to be a non-Archimedean superreal field, of which the surreal field is an instance. 

Benci et. al are concerned with giving an alternative definition of limit and an alternative formulation of Countable Additivity for the non-Archimedean probability theory. It would be interesting to see whether some of their ideas will also work for a surreal probability theory with Countable Additivity.\footnote{Although some philosophers following de Finetti reject Countable Additivity (CA), there are reasons to preserve something like CA in the theory. Many theorems for analysis on infinite sample spaces depend on CA. However, Benci et al.'s techniques ([2013] and [2016]) using Omega-limit to extend  CA  to an analogue in non-standard analysis (hyper-countable-additivity) give us hope that such an extension is possible in general. It would be interesting to explore the consequences of extending Omega-limit to the surreal field. We hope to write up in a second paper about surreal infinite sum and explore versions of CA on the basis of Benci et al's results.} 

However, in this paper we are concerned with decision theory with finite state spaces, so Finite Additivity is sufficient for our needs. The absence of Countable Additivity gives us much freedom in how we assign probabilities to infinite sets. We can, for example, assign probability $\frac{1}{\omega}$ to an infinite fair lottery on $\mathbb{N}$ and a probability $\frac{1}{2\omega}$ to an infinite fair lottery on $\mathbb{N}\backslash 2$. 
 Nonetheless, it is interesting to see that Benci et al.'s alternative axiomatization not only accommodates finitely-additive surreal probabilities but also forces the use of a non-Archimedean field in some cases.

\section{Pascal's Wager}

For a less contrived look at the advances surreal arithmetic allows us, we will use it to analyze one of the oldest problems using infinite utility: Pascal's Wager. Pascal's Wager makes a tantalizing offer: once one understands the incentive structure of the afterlife, there is a strong practical reason to live religiously. This is almost independent of any evidential considerations; so long as one has some credence that there is a god offering her followers an infinitely good afterlife, one ought to follow that god's religion. Thus, we are given a chance to do an end run around the evidence! As long as there is no proof of atheism, it is rational (perhaps rationally required) to behave as a convert. It thus occupies a special place in the pantheon of theistic arguments; it offers defeat to the atheist for mostly non-epistemic reasons. In the face of problems of evil and divine hiddenness, this is no small thing. 

Nearly as old as decision theory itself, the wager occupies an interesting place in the history of decision theory, as a problem involving (finitely many) infinite-value states. We will use it as a 'real world' test case for our surreal decision theory,  focusing its lens on this venerable argument and analyzing its most common twists and objections. 
 
Pascal argued\footnote{See Pascal [1670], section 418.} that the decision whether or not to lead a Christian life could be modeled as a decision problem with four states: either there was a god or not, and one either lives a Christian life or not. Three of these states have finite utilities: those in which there is no god, or in which there is a god but one is not Christian. The state in which there is a god and one is Christian, on the other hand, has infinite utility. Any finite gain or loss is swamped by the infinite value. Thus, Pascal reasoned, it is best to lead a Christian life as long as one's credence that there is a god is non-zero. The rule of expected utility maximization confirms this.

Many criticisms have been leveled against this argument.\footnote{Most notably H\'ajek 2003, but see Jordan [2007] for a thorough review.} We will pay special attention to two of them, since they display interesting features of our proposal. First, we will examine Alan H\'ajek's ``mixed strategy'' objection.  Next, we will examine the ``Many Gods'' objection. Our conclusion: while the move to surreal utilities blocks H\'ajek's objection (theological rejoinders notwithstanding), the Many Gods objection shows that the correct bet in a Pascalian decision problem depends crucially on an agent's credences, and therefore that the wager argument fails to deliver a non-epistemic argument for theism. 

\subsection{Mixed Strategies}

Typically, Pascal's wager has been set up as a decision problem with two options. 

\begin{table}[htbp]
\begin{center}
\begin{tabular}{ | l | r | r | r |}
\hline
  & God & No God & Expected Payoff\\ \hline                       
 Christian & $\infty$ & 10 & Infinite \\ \hline
 Non-Christian & 5 & 10 & Finite \\ \hline
\end{tabular}
\caption{Pascal`s Wager, Classical Presentation}
\end{center}
\end{table}
\noindent
But decision theorists know better. Whenever we have gambles, we can adopt mixtures of those gambles. We can think of mixtures heuristically as using coin flips to decide which gamble to take. So someone presented with the decision in Table 1 might make her choice by flipping a fair coin. That is, she has 0.5 chance of leading a Christian life and 0.5 chance of leading a non-Christian life. In that case, nonetheless, since a 0.5 chance of infinity is still infinite, the expected utility of the flip strategy = the expected utility of simply picking ``Christian.'' In fact, the coin can be arbitrarily biased against Christian, and still the mixed strategy has \emph{the same expected utility as the pure ``Christian'' option}! This is counterintuitive because gambles with arbitrary biases will have the same expected utility and the agent ought to be indifferent among the different gambles. As H\'ajek\footnote{H\'ajek [2003].} convincingly argues, if we assume the Principle of Regularity (the thesis that we should assign probability 1 only to logical truths and 0 only to contradictions), then any act should have a non-zero probability for the eventual outcome of becoming a Christian. Therefore, if we assume Regularity, any act can be seen as a mixed strategy of Pascal's Wager. This has the absurd consequence that we should be indifferent among all practical actions, thus trivializing practical reasoning and decision theory. Since practical reasoning and decision theory are not useless, there must be something wrong or mathematically incoherent with Pascal's Wager. So says the opponent.

We think, however, the real problem lies in the mathematical representation of Pascal's Wager. This is because $\infty$, at least in the extended reals and the more familiar Cantorian realms, has the absorption property.\footnote{In H\'ajek's terms:\textit{reflexive under multiplication.}} By standard lights, a chance at $\infty$ is as good as a sure-thing $\infty$. But in surreal arithmetic, this is not true. The surreal $\omega$ is strictly greater than the surreal $.5\omega$. Indeed, the surreal $\omega$ is strictly greater than any $p\omega$ for all $p\in(0,1)$. Thus, our proposal, representing the decision matrix with surreal numbers instead of extended reals, correctly predicts that the pure ``Christian'' strategy beats all mixed strategies.\footnote{ We note that our proposal is not the only one to do this. See Bartha [2007] and Herzberg [2011] for alternate proposals that make the same prediction. We do note that all of these proposals make use of non-Archimedean utilities, of which we shall say more soon.}

H\'ajek noted the potential for surreal valued utilities to escape his objection.\footnote{H\'ajek [2003].} Instead, he argues that surreal infinite numbers do not have the same properties as the infinity Pascal seems to be talking about. H\'ajek's Pascal sees salvation as the greatest good, and thus possessing the absorption property for addition. We do not dispute H\'ajek's reading of Pascal\footnote{However, it appears that H\'ajek  no longer accepts this reading of Pascal.} (although we express some skepticism about the theology underlying a view according to which salvation is the greatest good, or indeed the rationality of a view whereby salvation and no apple is as preferable as salvation plus an apple),\footnote{But see Herzberg [2011] for discussion, especially for a hyperreal model that satisfies both reflexivity under addition and the non-reflexivity under multiplication.} but we are less interested in giving a faithful representation of Pascal's original argument than we are in applying our more general proposal to this problem in transfinite decision theory. Indeed, we think that Pascal falls to another objection.

\subsection{Many Gods}

A common objection to Pascal is that his decision problem is too simple, and as a result, the use of infinite utilities looks less problematic than it is.\footnote{The objection is as old as Diderot [1746], but has received a more rigorous formulatin in e.g. Cargile [1966]. Herzberg [2011] briefly discusses this possibility but focuses instead on modeling agents who do not countenance it.} For there are a great many purported gods, many of which treat their followers well, and their doubters cruelly. Moreover, there are any number of other potential eschatological situations. Perhaps there is a god, but god is a universalist, so that everyone ends well. Perhaps there is a god, but god is a rationalist, and so anyone who makes epistemic decisions (like belief in a god) for pragmatic reasons ends poorly. The objection goes that once we see all these situations, and their accompanying infinite utilities and disutilities in the decision problem, we conclude that there's nothing interesting to say, and so problems of this sort aren't sensibly posed.\footnote{Rescher [1985] presses this line of reasoning, although not explicitly in connection with the many gods objection.}

What would our surreal decision theory predict? Our theory, it turns out, allows us to formulate and analyze the Many Gods objection in a precise way. Let $E_1...E_n$ be a partition over states in an expanded Pascalian decision problem. With each $E_i$, we associate some surreal number $n$, corresponding to $u(E_i)$ in the agent's utility function. Let $cr(E_i)$ be value of the agent's credence function over $E_i$. We can then give the EU of each of the $E_i$'s. 

For example, suppose our agent thinks that there are three live divine candidates: Zeus, Apollo and Athena. She then has four religious options: Zeusianism, Athenianism, Apollinism, and Atheism. Zeusianism is an exclusivist religion. Zeusians get infnite utiity, but everyone else is damned. But according to Athenian theology, Athena is a universalist who will give everyone infinite utility. According to Apollinism, Apollo rewards atheists and damns everyone else.\footnote{Such Apollo is thus an example of the sort of god no one believes, but is regularly trotted out by philosophers objecting to Wager-style arguments. We propose Silly Theism as a technical name for this type of religion.} We may represent the problem in the following table, representing the finite utility of a regular life as 100:\\

\begin{table}[htbp]
\begin{center}
\begin{tabular}{ | l | c | c | c | r | }
  \hline                       
  & Zeus & Athena & Apollo & Atheism\\ \hline
  Zeusian & $\omega$ & $\omega$ & -$\omega$ & 100 \\ \hline
  Athenian & -$\omega$ & $\omega$ & -$\omega$ & 100 \\ \hline
  Apollinist & -$\omega$ & $\omega$ & -$\omega$ & 100 \\ \hline  
Atheist & -$\omega$ & $\omega$ & $\omega$ & 100 \\
  \hline  
\end{tabular}
\caption{Pascal's Wager With Three Gods}
\end{center}
\end{table}
Already, using surreal values allows her to assign a sensible ranking of her options. Which religion is best will depend on what her credence function is like. If, for instance, Cr(Zeus) = .5, Cr(Athena) = .3, Cr(Apollo) = .1, and Cr(Atheism) = .1, then: 
\begin{eqnarray}
EU(Zeusian) & = & .7\omega + 10 \nonumber \\ &>& -.1\omega + 10 = EU(Atheist)  \nonumber \\ &>&  -.3\omega + 10 =EU(Athenian) = EU(Apollinist)  \nonumber
\end{eqnarray}
With the credence favoring Zeus, Zeus is the best option. On the other hand, if Cr(Zeus) = .1, Cr(Athena) = .2, Cr(Apollo) = .2, and Cr(Atheism) = .5, then:
\begin{eqnarray}
EU(Atheist) & = & .3\omega + 50 \nonumber \\ &>& .1\omega + 50 =EU(Zeusian) \nonumber \\ &>&  -.1\omega + 50 =EU(Athenian) = EU(Apollinist) \nonumber
\end{eqnarray}
With the credence favoring atheism, Atheist is the best option.\footnote{As we have set things up, the exclusivist and atheist options each dominate the universalist and silly options, but other combinations (such as scenarios with multiple exclusivist gods in play, or mildly inclusivist options where some gods favor some infidels over others) can bring out the benefits of those.} 

A further complication is that infinite utilities come in different degrees of magnitudes. We can represent this in the decision matrix by allowing the utility function to range over the entire hierarchy of infinities, with the subscript indexing their degrees: 

\begin{table}[htbp]
\begin{center}
\begin{tabular}{ | l | c | c | c | r | }
  \hline                       
  & Zeus & Athena & Apollo & Atheism\\ \hline
  Zeusian & $\omega_{100}$ & $\omega_0$ & -$\omega_{5}$ & 100 \\ \hline
  Athenian & -$\omega_{100}$ & $\omega_0$ & -$\omega_{5}$ & 100 \\ \hline
  Apollinist & -$\omega_{100}$ & $\omega_0$ & -$\omega_{5}$ & 100 \\ \hline  
Atheist & -$\omega_{100}$ & $\omega_0$ & $\omega_{137}$ & 100 \\
  \hline  
\end{tabular}
\caption{Pascal's Wager With a Hierarchy of Infinities}
\end{center}
\end{table}

Again, which religion is best will depend on what her credence function is like.

Take the first example: since Cr(Zeus) = .5, Cr(Athena) = .3, Cr(Apollo) = .1, and Cr(Atheism) = .1,  
\begin{eqnarray}
EU(Zeusian) & = & .5\omega_{100} + .3\omega_{0} -0.1\omega_{5} + 10 \nonumber \\ &<&-.5\omega_{100} + .3\omega_{0} +0.1\omega_{137} + 10 = EU(Atheist) \nonumber
\end{eqnarray}
Since $\omega_{137}$ is larger than any finite product of the lower infinities, Atheist is the best option. 

On the other hand, if we allow the credence function to range over surreal infinitesimal values, then things can be very different. Let Cr(Zeus) = .5, Cr(Athena) = .3, Cr(Apollo) = $\frac{1}{\omega_{137}}$, and Cr(Atheism) = $0.2 - \frac{1}{\omega_{137}}$. Then: 
\begin{eqnarray}
EU(Zeusian) & = & .5\omega_{100} + .3\omega_{0} -\omega_{5} / \omega_{137} + 20 - 100/\omega_{137} \nonumber \\ &>&-.5\omega_{100} + .3\omega_{0} - 1 + 20 - 100/\omega_{137} = EU(Atheist) \nonumber
\end{eqnarray}
Since $\omega_{137}$ is in the denominator, $\omega_{100}$ is the dominating infinity here, and Zeusianism is the best option.

We can do this for arbitrarily complicated decision problems of this sort. So there is nothing incoherent or problematic about the use of infinite utilities. But the argument does not deliver the result as advertised. Its partisans sell Pascal's Wager as a route to religion that does not depend on how the evidence falls between the theist and atheist. Our theory says otherwise. What one should do in a Pascalian decision problem depends on what one's credence function is like. It is not, after all, an end run around the evidence. For one's credence function should be sensitive to different kinds of evidence that support competing hypotheses to different degrees. As a result, one's expected utility function will vary accordingly, giving different answers to the question: ``what ought one rationally to do?'' 

\subsection{Why Not Finite Utilities?}

The analysis we have given of the Many Gods objection is very similar to that of Mougin and Sober [1994], in particular their finite-utility model (where the utility of salvation is very great, but not infinite). So one might justly ask: what does Surreal Decision Theory have to offer that can't be had using finite numbers? 

It is true that surreal arithmetic and finite arithmetic have many similar properties. This is by design. One of the great virtues of using surreal numbers in decision theory is the fact that surreal arithmetic treats finite and infinite numbers alike, and in a way that extends finite arithmetic to include infinite numbers. Unlike more Cantorian arithmetics, it lacks the absorption property and other features that make up the familiar differences between finite and infinite arithmetic. And so it is unsurprising that a surreal analysis of the Many Gods objection will look similar to a finitist analysis. The arithmetic is similar, and since the move to surreal utilities is primarily a technical change, the philosophical issues are similar. However, the difference in field structure between \textbf{No} and $\mathbb{R}$ means that there are Pascalian problems which can be modeled using surreal numbers that cannot be using real numbers. 

Consider, for instance, an agent, Theresa, trying to decide between two gods, each offering an $\omega$-valued afterlife. Her credence that Odin is the true god is .6, and that Ra is the true god is .4, so she tentatively elects to wager for Odinism. Moreover, even though her utility is linear in dollars, there is no amount of money that the Cult of Ra could pay her to change her wager. Theresa values salvation infinitely with respect to earthly goods. Surreal decision theory can give a model of her preferences; finitist decision theory cannot, for whatever number we assign the utility of salvation, by the Archimedean axiom there is some finite amount of money the Cult of Ra could pay her to make conversion rational.  Consequently, surreal decision theory will be able to model agents in Pascalian decision problems that finite decision theory cannot: namely, those who cannot be paid to wager against the evidence. Or, more generally, those whose preference for salvation over earthly goods is genuinely non-Archimedean.\footnote{There might be an analogous issue in surreal decision theory. If we allow the difference between Theresa's credence in Odinism and her credence in Raism to take a small enough (how much counts as `enough' depends on the uitlity of salvatiuon) infinitesimal value, then the Cult of Ra could bribe her to join. This follows from our Continuity$\star$ axiom. We acknowledge this as a limitation of surreal decision theory; it shows a kind of discontinuous preference structure that we cannot capture. But it does not represent a case that finite utility theory handles better than we do. Thus, we claim superiority for our theory over finite utility theory and naive infinite utility theory, even if we cannot capture everything we might like to.}

We assumed Theresa's utility is linear in dollars. Even if no human agent has a psychology like this, it's not an implausible psychology; there could well be or have been agents with it who try to decide what religion to follow, and a decision theory should be able to give good models of their deliberations. Moreover, any attempt to give a finitist reinterpretation of Theresa's preferences using the declining marginal utility of money will face an issue similar to attempts to reinterpret the Allais preferences pointed out by Buchak [2014]: it will imply other odd preferences and tradeoffs between goods and money. For example: let n be the value of salvation. In the proposed finitist  reinterpretation, Theresa heavily discounts large dollar amounts. In the example, no amount of money is equal to the utility provided by $.2n$, which is the difference in expected value between Odinism and the Cult of Ra. 

Of course, .2n may still be a large amount of utility. But we can make it smaller by adjusting the credences. If Theresa's preference for salvation over money is truly infinite, then even a small evidential advantage for Odinism will be good enough to make it impossible to pay her to convert. So suppose cr(Odinism) - cr(Cult of Ra) = $1/n$, where n is the utility assigned to salvation. In the revised case, the finitist must interpret her as valuing no amount of money more than a single utile. And with further adjustment, we can make the value of all the money in the world arbitrarily small. This is, to say the least, as implausible a psychology as one where money is linear in utility. 

\subsection{Degrees of Glory}

Before concluding this section, we would like to point out two related features of surreal decision theory. 

First, in many theological traditions, there exist different ``degrees of glory'' of the afterlives, corresponding to different kinds of behaviors during the earthly life. We find variants of this doctrine in 
 Buddhism,\footnote{For example, the Theravada School of Buddhism holds that there is a cycle of rebirth, and the condition of someone's rebirth (the next life) depends on that person's earthly deeds.} 
Judaism,\footnote{See the Jewish Merkavah and Heichalot literature for a detailed discussion of the seven heavens.}
 Christianity,\footnote{``But who can conceive, not to say describe, what degrees of honour and glory shall be awarded to the various degrees of merit? Yet it cannot be doubted that there shall be degrees.'' (St. Augustine, \emph{The City of God}, 22: 30. }
Islam,\footnote{The concept of the seven heavens occurs in the Qur'an (41:12, 65:12, 71:15) as well as in the hadiths.}
and Mormonism.\footnote{``Paul ascended into the third heavens, and he could understand the three principal rounds of Jacob's ladder--the telestial, the terrestrial, and the celestial glories or kingdoms, where Paul saw and heard things which were not lawful for him to utter. I could explain a hundred fold more than I ever have of the glories of the kingdoms manifested to me in the vision, were I permitted, and were the people prepared to receive them.''
(Joseph Smith, \emph{History of the Church}, 5:402).}

We can model these ``degree of glory'' eschatologies by further complicating our matrix. Although those traditions suggest only different heavenly rewards, not different infinities, we will follow Pascal in modeling ``good'' religious afterlives as infinite in value compared to earthly goods. This will allow us to showcase another feature of surreal decision theory: the ability to model preferences among options that are themselves all of some infinite value or other. 

As a simplifying assumption, we'll assume that gods who have a degree of glory policy reward better degrees to their better followers, and worse degrees to their worse followers. Thus, with two gods in play we have four outcomes: God 1 exists and I am good, God 1 exists and I am bad, God 2 exists and I am good, God 2 exists and I am bad. We also assume that the various gods are exclusivists and offer penalties and rewards on a par with each other. But it should not be difficult to construct a (more complicated) model truer to the source texts and including the other twists we've discussed. 
\begin{table}[htbp]
\begin{center}
\begin{tabular}{ | l | c | c | c | c | r | }
  \hline                       
  & Zeus \& Good & Zeus \& Bad & Athena \& Good & Athena \& Bad & Atheism\\ \hline
  Zeusian & $\omega^{2}$ & $\omega$ & -$\omega$ & -$\omega$ & 100 \\ \hline
  Athenian & -$\omega$ & -$\omega$ & $\omega^{2}$ & $\omega$ & 100 \\ \hline  
Atheist & -$\omega$ & -$\omega$ & -$\omega$ & -$\omega$ & 100 \\
  \hline  
\end{tabular}
\caption{Pascal's Wager With Degrees of Glory}
\end{center}
\end{table}

The matrix above shows a very simple ``degrees of glory'' decision, but it allows us to illustrate an interesting dimension that these types of theology can add to the problem. For certain credence functions, even if the balance of our credence fails in favor of the existence of God 1, following God 2 might still be more rational. This will happen when we are fairly confident that we will be a good follower of God 2, but a bad follower of God 1. Taking the utilities as above, we can see that Cr(Zeus) = Cr(Zeus and Good) + Cr(Zeus and Bad), and similarly for Cr(Athena). Thus, if Cr(Zeus and Good) = .1, Cr(Zeus and Bad) = .5, Cr(Athena and Good) = .3, and Cr(Athena and Bad) = .1, our credence favors Zeus, but the expected utility of Zeusianism = (.1$\omega^2 + .5\omega - .4\omega$), while the expected utility of Athenianism = (.3$\omega^2 + .1\omega - .6\omega$). This does not undermine our claim that credence is relevant to the right decision in a Pascalian problem. But it does show that it can be rational to bet against the most likely option in interesting cases. 

We would like to emphasize another feature of surreal decision theory. Table 3 and Table 4 might suggest the existence of an ever-greater hierarchy of infinite utilities. This leads to a worry raised by H\'ajek:
\begin{quote}
[T]he chosen utility for salvation, in turn, is not merely bettered, but swamped to the same degree by another conceivable utility: for instance, $\omega^2$ stands to $\omega$ as $\omega$ stands to $1$ and so on. And that other utility, in turn, is swamped by still another ($\omega^3$, say), and so on ad infinitum—an ‘infinitum’ of the form that Pascal would recognize! Far from being the best possible thing, salvation isn’t even close; in fact, in the eyes of Pascal it becomes a pure nothing. It is hardly surprising, then, that the notion of infinity that he envisages is reflexive under addition. At least that way infinitude stays infinite-looking.\footnote{H\'ajek [2003], pp. 46-47.}
\end{quote}
Implicit in H\'ajek's critique is the thought that any number in the codomain of the utility function is a ``conceivable'' utility. We reject this as overly realistic about the numbers used in utility representation. It is important to recall that the utility functions are only unique up to linear transformations. Thus, we should not take the actual numbers used in an expected utility representation of an agent very seriously. We can make the number assigned to a given good arbitrarily high or low. What's important is the underlying ordering. Thus, if salvation is of infinite value compared to apples, then whatever utility we give to an apple, we must give salvation one that is infinitely bigger. We could, if we wish, assign an apple the utility of $\omega$; we would just then be required to assign salvation a much bigger infinity as its utility. For example, preferences with a representation where $u$(apple) = 1 and $u$(salvation) = $\omega$ are just as faithfully represented by one where $u$(apple) = $\omega$ and $u$(salvation) = $\omega^2$. Even if nothing in the original representation was assigned $\omega^2$ as its utility. 

The point: where H\'ajek sees conceivable utilities, we see possible representations of utilities. But the fact that we could represent something's utility with a number that is infinite with respect to the number we have chosen to represent the utility of salvation does not imply that there is conceivably something with utility that is infinite with respect to that of salvation. For that, we would need something in our space of lotteries that is infinitely preferable to salvation. And if there is such a thing, no specifics of the number system chosen as the codomain of the utility function will prevent salvation from being swamped. Any faithful representation will swamp it.

\section{Conclusion}

We set out to make sense of transfinite decision theory: the study of decision problems involving infinite utilities. It faces well-known problems, violating many of our strong intuitions such as the dominance principle, and failing to deliver well-defined answers to seemingly sensible questions. 

We propose a solution: better technology---John Conway's surreal numbers. Because the surreals form a universally embedding ordered field, they include all finite, infinite, and infinitesimal numbers. Because their associated arithmetic operations are commutative and non-absorptive, the problems with infinite utilities in finite state spaces evaporate. We have not addressed the problems unique to infinite state spaces (such Alan H\'ajek and Harris Nover's Pasadena Game), but it is already work in progress and we hope that our framework will provide a viable treatment pending ongoing research in surreal analysis. 


Applying our theory to Pascal's Wager, we provide precise formulations to two well-known objections: Mixed Strategies and Many Gods. Our theory correctly predicts that the pure ``Christian'' strategy beats all mixed strategies but the ultimate normative verdict depends crucially on one's credence function. We also provide the first treatment of ``degrees of glory" theologies in a formal, decision-theoretic framework. In so doing, we fend off one of H\'ajek's objections to a surreal analysis of the wager argument. Nevertheless, we conclude that Pascal's Wager fails as a purely pragmatic argument for adopting a religious life. Credence cannot be cut out of the equation.

In closing, we note future potential applications for surreal numbers. We see our project as the first step in a program of bringing cutting-edge mathematical tools to bear on old philosophical problems. We expect the use of surreals to be particularly helpful in solving problems in transfinite axiology, infinite physical quantities,\footnote{Such as those in quantum field theory and thermodynamics, the latter of which has been investigated by Philip Ehrlich (1982).} and in dissolving many of the traditional paradoxes of infinity, which rely on the shortcomings of standard real analysis and cardinal arithmetic. 

\section*{Acknowledgement}
We thank the editors and an anonymous reviewer at PPR for their constructive comments, which led to much improvement of the paper. 
We are grateful for discussions with 
Amanda Askell,
David Black,
Paul Bartha, 
Liam Kofi Bright,
Catrin Campbell-Moore,
John Conway, 
Nicholas DiBella,
Kenny Easwaran,
Andy Egan,
Adam Elga, 
Branden Fitelson,
Simon Goldstein, 
Remco Heesen,
Leon Horsten,
Liz Jackson,
Pavel Janda,
Mark Johnston, 
Xiaofei Liu, 
Yang Liu,
Sarah Moss,
Richard Pettigrew,  
Patricia Rich, the Rutgers Formal Epistemology Reading Group, and especially Philip Ehrlich and Alan H\'ajek. 

\appendix
\section{Appendix}

Here we present a full proof of the surreal version of the von Neumann-Morgenstern representation theorem. 

\noindent
{\sc Notation 1}:  Let  $\star$ denote the natural embedding from the standard universe into the surreal universe. Let  \textbf{No} denote a surreal model. \\
\\
\noindent
{\sc theorem 1 (Surreal von Neumann-Morgenstern Theorem)}: Let X be a finite space of lotteries, and let $\preceq$ be a binary relation $\subseteq X \times X$. Then $\preceq$ admits an expected utility representation $U : X \rightarrow \textbf{No}$ such that $\forall x, y \in X$\\

$U(x) \leq U(y) \Leftrightarrow x \preceq y$ if and only if $\preceq$ satisfies all of the following:
\begin{enumerate}
\item Completeness: $\forall x, y \in X,$ either $x \preceq y$ or $y \preceq x$.
\item Transitivity: $\forall x, y, z \in X,$ if $x \preceq y$ and $y \preceq z$, then $x \preceq z$.
\item Continuity$\star$: $\forall x, y, z \in X,$ if $x \preceq y \preceq z$, then there exist a surreal $p \in \star[0, 1]$ such that $px + (1 - p)z \sim y$.
\item Independence$\star$: $\forall x, y, z \in X, \forall p \in \star(0, 1],$ $x \preceq y$ if and only if $px + (1 - p)z \preceq py + (1 - p)z$.
\end{enumerate}

\textit{Proof:} We adopt the usual constructive proof strategy for the von Neumann-Morgenstern representation theorem. We will use the proof to illustrate the content of Continuity$\star$ and Independence$\star$ as well as some properties of surreal numbers. 

($\Rightarrow$) This is, as usual, the easier direction. Suppose the existence of  an expected utility representation $U : X \rightarrow \textbf{No}$ such that $\forall x, y \in X$, $U(x) \leq U(y) \Leftrightarrow x \preceq y$. We want to show that $\preceq$ satisfies Completeness, Transitivity, Continuity$\star$ and Independence$\star$. 

(Completeness) Take any $x, y \in X$, suppose that it is not the case that $x \preceq y$. Then it is not the case that $U(x) \leq U(y)$. Since $U(x), U(y) \in \textbf{No}$, $U(x) \geq U(y)$. (Application of a theorem about  $\leq$ as a linear ordering of the surreal field.) Thus, $y \preceq x$. 

(Transitivity) Take any $x, y, z \in X$, suppose that  $x \preceq y$ and $y \preceq z$. Then $U(x) \leq U(y)$ and $U(y) \leq U(z)$. Since $U(x), U(y), U(z) \in \textbf{No}$, $U(x) \geq U(z)$. (Application of a theorem about  $\leq$ as a linear ordering of the surreal field.) Thus, $x \preceq z$. 

(Continuity$\star$) Take any $x, y, z \in X$, suppose that  $x \preceq y \preceq z$. Then $U(x) \leq U(y) \leq U(z)$. Now, $U(x), U(y), U(z) \in \textbf{No}$. Since in \textbf{No} infinitesimals are well-defined, take $p = \frac{U(y)-U(z)}{U(x)-U(z)}$. Then, $pU(x) + (1 - p)U(z) = U(y)$. By the well-known fact that any expected utility representation is linear, we have that $px + (1 - p)z \sim y$.

(Independence$\star$) Take any $x, y, z \in X, p \in \star(0, 1].$ We have: 
\begin{align*}
x \preceq y &\Leftrightarrow U(x) \leq U(y) \\
 &\Leftrightarrow pU(x) \leq pU(y) \\
 & \Leftrightarrow pU(x) + (1-p)U(z) \leq pU(y) + (1-p)U(z) \\
 &  \Leftrightarrow  px + (1 - p)z \preceq py + (1 - p)z
\end{align*}

($\Leftarrow$) Suppose that $\preceq$ satisfies Completeness, Transitivity, Continuity$\star$ and Independence$\star$. We want to construct a $\star$-affine function $U : X \rightarrow \textbf{No}$ such that $\forall x, y \in X, U(x) \leq U(y) \Leftrightarrow x \preceq y$. As usual\footnote{The following proof follows closely Jonathan Levin's online notes at: http://web.stanford.edu/~jdlevin/Econ\%20202/Uncertainty.pdf. Accessed on March 7, 2015. }, let $\overline{p}  $ and $\underline{p}$ denote the $\preceq$-top and $\preceq$-bottom elements in $X$. If $\preceq$ admits several maximals and several minimals, then let $\overline{p}  $ and $\underline{p}$ denote some representatives of the equivalence classes of maximals / minimals. If $\overline{p} \sim \underline{p}$, then choose any constant surreal function and we are done. Suppose $\overline{p} \succ \underline{p}$.  By Continuity$\star$ and Independence$\star$, suppose that $1>b>a>0$, we have: 
\begin{align*}
\overline{p} & \sim b\overline{p} + (1-b)\overline{p} \\ 
&\succ b\overline{p} + (1-b)\underline{p} \\
&\sim  (b-a)\overline{p} + a\overline{p}+ (1-b)\underline{p} \\
&\succ  (b-a)\underline{p} + a\overline{p}+ (1-b)\underline{p} \\
&\sim   a\overline{p}+ (1-a)\underline{p} \\
&\succ \underline{p}
\end{align*}
Thus, 
\begin{equation}
\overline{p} \succ  b\overline{p} + (1-b)\underline{p} \succ   a\overline{p}+ (1-a)\underline{p} \succ \underline{p} 
\end{equation} 

(Lemma) $\forall p \in X, \exists ! \lambda_p \in $ \textbf{No} such that $\lambda_p \overline{p} + (1-\lambda_p) \underline{p} \sim p$.\\

Existence:  By Continuity$\star$, for $\overline{p} \succeq p \succeq \underline{p}$, there is a surreal $\lambda_p$ s.t. $\lambda_p \overline{p} + (1-\lambda_p) \underline{p} \sim p$. 

Uniqueness: By Inequality (1), if there are $\lambda_1$ and $\lambda_2$ s.t. $\lambda_1 > \lambda_2$, then $\lambda_1 \overline{p} + (1-\lambda_1) \underline{p} \succ \lambda_2 \overline{p} + (1-\lambda_2) \underline{p}$. Thus, there is at most one $\lambda_p$ s.t. $\lambda_p \overline{p} + (1-\lambda_p) \underline{p} \sim p$. 

Therefore, (Lemma) is true. 

Now, because of (Lemma), we can construct the desired utility function as $U(p) = \lambda_p$. We know: 
$$p \succeq q \Leftrightarrow \lambda_p \overline{p} + (1-\lambda_p) \underline{p}  \geq \lambda_q \overline{p} + (1-\lambda_q) \underline{p} \Leftrightarrow \lambda_p \geq \lambda_q.$$

In the final step of the proof, we show that $U$ is linear, i.e. $$\forall a \in \star[0,1], \forall p, p' \in X, U(ap + (1-a)p') = aU(p) + (1-a)U(p'). $$

Take $a\in \star[0,1], p, p' \in X$. By the construction of $U(p)$, we have: $p \sim U(p) \overline{p} + (1-U(p)) \underline{p}$ and $p' \sim U(p') \overline{p} + (1-U(p')) \underline{p}.$ Thus, 
\begin{equation}
ap+(1-a)p' \sim (aU(p)+(1-a)U(p'))\overline{p} + (1-(aU(p)+(1-a)U(p'))) \underline{p}
\end{equation}

By the construction of $U(p)$, we know that $U(ap + (1-a)p')$ is the unique $\lambda$ s.t. $ap+(1-a)p' \sim \lambda \overline{p} + (1-\lambda) \underline{p}.$ Because of (2), $\lambda = aU(p) + (1-a)U(p').$ 

 Therefore, $U(ap + (1-a)p') = aU(p) + (1-a)U(p')$. So $U$ is indeed a linear utility function. Moreover, any linear utility function has an expected utility form (since any lottery $x$ can be written as a probabilistic mixture of lotteries in $X$). So $U$ has an expected utility form.  $\Box$\\

\newpage
Works Cited
\begin{enumerate}
	\begin{footnotesize}
		\item Augustine [1993]. \emph{The City of God}, trans. Marcus Dods. New York: Random House, Inc. 
		\item Bartha, Paul [2007]. ``Taking Stock of Infinite Value: Pascal's Wager and Relative Utilities." \textit{Synthese}, 154: 5-52.
		\item Bartha, Paul [MS]. ``Making Do Without Expectations.'' 
		\item Benci, Vieri, Leon Horsten, and Sylvia Wenmackers [2013]. ``Non-Archimedean Probability.'' \emph{Milan Journal of Mathematics}, \textbf{81}, pp. 121-51. 
		\item Benci, Vieri, Leon Horsten, and Sylvia Wenmackers [2016]. ``Infinitesimal Probabilities.'' \emph{British Journal for the Philosophy of Science}, advance access. 
		\item Buchak, Lara [2014]. \emph{Risk and Rationality}, Oxford University Press. 
		\item Cargile, James [1966]. ``Pascal's Wager," \textit{Philosophy}, 35: 250-7.
		\item Colyvan, Mark [2006]. ``No Expectations," \textit{Mind}, 115:695-702.
		\item --- [2008]. ``Relative Expectation Theory" \textit{Journal of Philosophy}, 105(1): 37-44.
		\item Colyvan, Mark, and Alan H\'ajek [2016]. ``Making Ado Without Expectations.'' \emph{Mind}, published online May 6, 2016. 
		\item Conway, John [1974]. \textit{On Numbers and Games}, Natick: A.K. Peters/CRC Press.
		\item Diderot, Denis [1746]. \textit{Pensees Philosophiques}, reprinted Whitefish, MN: Kessinger Publishing, 2009.
		\item Easwaran, Kenny [Ms.]. ``Dr. Truthlove, or How I Learned to Stop Worrying and Love Bayesian Probabilities."
		\item Ehrlich, Phillip [1982]. ``Negative, Infinite, and Hotter than Infinite Temperatures," \textit{The Bulletin of Symbolic Logic} 18 (1): 1-45.
		
		\item Ehrlich, Phillip [2012]. ``The Absolute Arithmetic Continuum and The Unification of All Numbers Great and Small," \textit{Synthese} 50 (2): 233-277.
		\item Eriksson, Lina and Alan H\'ajek [2007]. ``What Are Degrees of Belief?" \textit{Studia Logica} 86 (2): 183–213.
		\item Fine, Terrence [2008]. ``Evaluating the Pasadena, Altadena, and St. Petersburg Gambles," \textit{Mind} 117:613-632
		\item  Gonshor, Harry [1986]. \emph{An Introduction to the Theory of Surreal Numbers}. London Mathematical Society, Lecture Note Series 110. Cambridge University Press.
		\item H\'ajek, Alan [2003]. ``Waging War on Pascal's Wager," \textit{Philosophical Review}, 113:27-56.
		\item --- [Forthcoming]. ``Unexpected Expectations," \textit{Mind}.
		\item H\'ajek, Alan and Harris Nover [2004]. ``Vexing Expectations," \textit{Mind} 113:237-249.
		\item --- [2006]. ``Perplexing Expectations," \textit{Mind} 115:703-720.
		\item --- [2008]. ``Complex Expectations," \textit{Mind} 117:643-664.
		\item Herzberg, Frederik [2011]. ``Hyperreal Utilities and Pascal's Wager." \textit{Logique Et Analyse} 213:69-108. 
		\item Jordan, Jeffrey [2007]. ``Pascal's Wager and James's Will To Believe," \textit{Oxford Handbook of Philosophy of Religion}, Oxford: OUP.
		\item Keisler, H. J. [1976]. \textit{Foundations of Infinitesimal Calculus}, Boston: Prindle, Weber \& Schmidt.
		\item Meacham, Christopher J. G. and Jonathan Weisberg [2010], ``Representation Theorems and the Foundations of Decision Theory." \textit{Australasian Journal of Philosophy}, 89 (4):641-663.
		\item Mougin, Gregory, and Elliott Sober [1994], ``Betting Against Pascal's Wager”, \emph{No\^us}, XXVIII: 382–395.
		\item Pascal, Blaise, [1670], \textit{Pensées}, translated by A. J. Krailsheimer. New York: Pengiun, 1966.
		\item \emph{The Qu'ran} [2010]. Tr. M. A. S. Abdel Haleem. Oxford: OUP.
		\item Ramsey, Frank P. [1926], ``Truth and Probability,'' in Ramsey [1931], \emph{The Foundations of Mathematics and other Logical Essays,} Ch. VII, p.156-198, edited by R.B. Braithwaite, London: Kegan, Paul, Trench, Trubner \& Co., New York: Harcourt, Brace and Company. 
		\item Rescher, Nicholas [1985]. \textit{Pascal's Wager}, South Bend, IN: Notre Dame University Press.
		\item Roberts, B. H. [1909]. \emph{History of the Church of Jesus Christ of Latter-day Saints}, volume 5. Salt Lake City: Deseret News.
		\item Tondering, Claus [2001]. ``Surreal Numbers: An Introduction,'' Version 1.6,\\
		http://www.tondering.dk/claus/surreal.html. 
		\item  Wenmackers, Sylvia, and Leon Horsten [2013]. ``Fair Infinite Lotteries.'' \emph{Synthese}, \textbf{190}, pp. 37-61. 
	\end{footnotesize}
\end{enumerate}

\end{document}